\documentclass[letterpaper, 10 pt,conference]{ieeeconf}
\IEEEoverridecommandlockouts    

\usepackage{multirow}
\usepackage{lipsum}
\usepackage{amsmath}
\usepackage{amssymb}
\usepackage[ruled,vlined]{algorithm2e}
\usepackage{graphicx}
\graphicspath{{./Figures/}}
\DeclareMathOperator*{\argmax}{arg\,max}

\title{\LARGE \bf Momentum Based Reward Design for Low Emission Traffic Signal Control}

\author{
	\parbox{\textwidth}{%
		\centering
		Chinmay Mundane$^{1}$, Amith Manoharan$^{1}$, Arun Singh$^{1}$%
	}%
	\thanks{$^{1}$Institute of Technology, University of Tartu,
		{\tt\small mundane@ut.ee, amith.manoharan@ut.ee, arun.singh@ut.ee}}%
	\thanks{This research was co-funded by the European Union and the Estonian Research Council through the  TEM-TA101 project. The work was also funded through the Estonian Research Council Grant PSG753.}
}

\begin{document}
	\maketitle
	\thispagestyle{empty}
	\pagestyle{empty}
	\begin{abstract}
    Urban traffic congestion is a growing global issue
    contributing significantly to long commute times
    and environmental pollution. Traditional traffic signal control systems often fail to adapt to dynamic traffic conditions. Adaptive traffic signal control can improve urban traffic without changing road infrastructure. Deep Reinforcement Learning (DRL) has shown strong performance for this task, but existing delay and queue-based rewards often produce short-sighted or unstable policies.

    This paper proposes a Momentum-Based Reward Function (MBRF) that encourages vehicles to keep moving rather than penalizing congestion alone. The method is evaluated in SUMO (Simulation of Urban MObility) using standard traffic metrics such as waiting time, queue length, throughput, and CO$_2$ emissions. Results show that the proposed reward produces better throughput-emission trade-offs and more stable learning behavior than delay or queue-based rewards, as well as classical controllers such as Max Pressure and LQF.

    \end{abstract}
	
	\section{Introduction}
	\label{sec:introduction}

\begin{figure}[t]
    \centering
    \includegraphics[width=\linewidth]{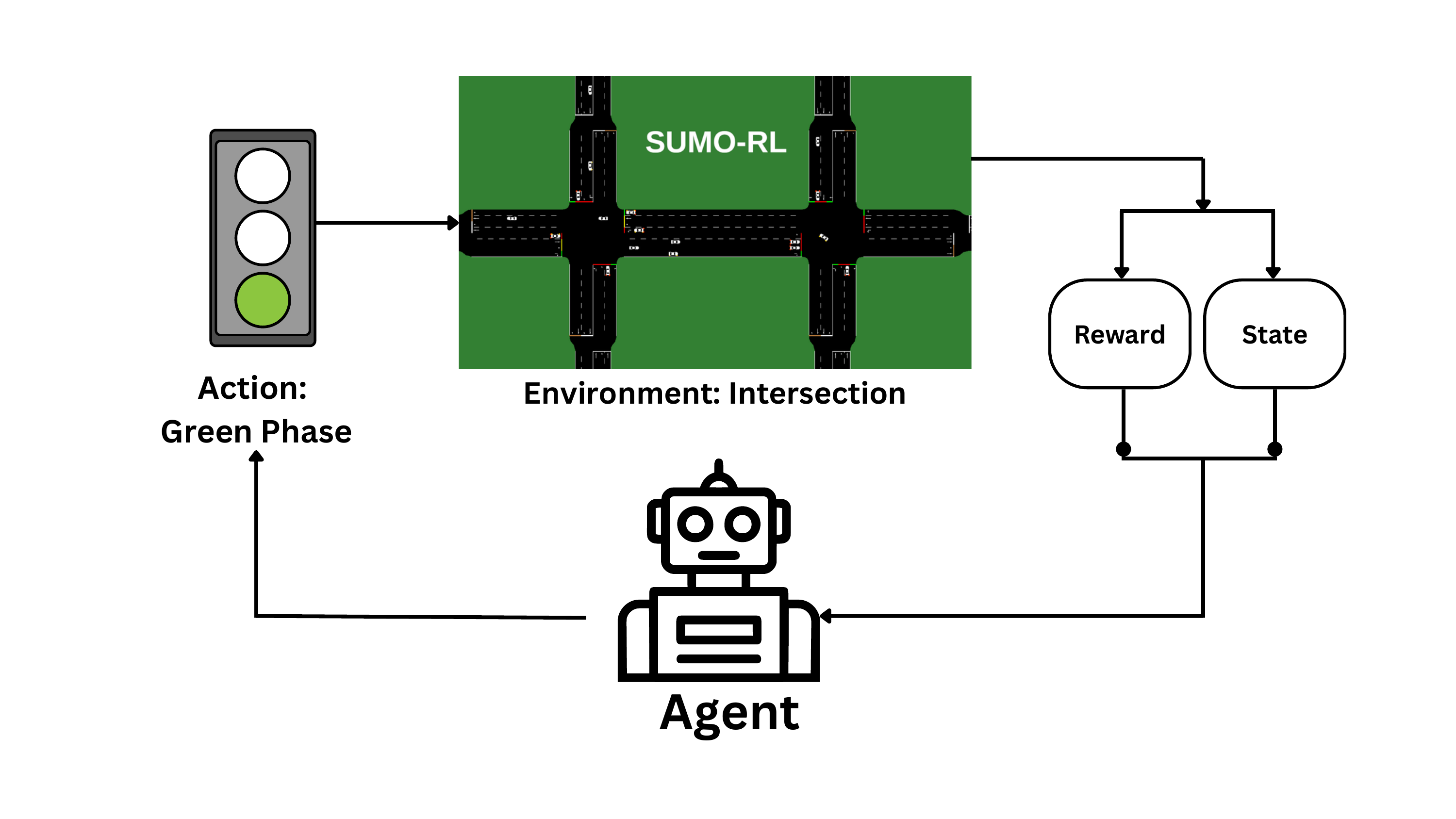}
    \caption{The traffic signal control problem is modeled as a Markov Decision Process (MDP) where the agent observes real-time intersection states (phase, density, queues) from SUMO via its programming interface (TraCI), selects signal phases as actions, and receives a momentum-based reward.}
    \label{fig:component}
    \vspace{-0.6cm}
    \end{figure}

    Urban traffic congestion imposes substantial economic and environmental costs, with recent analyses estimating billions in annual losses due to delayed travel times and lost productivity \cite{1.1, 1.2}. This motivates the need for intelligent control strategies that optimize existing infrastructure without major physical expansions. Traditional Adaptive systems (e.g., SCOOT \cite{scoot} and SCATS or LQF \cite{lqf}) rely on rule-based or model-driven logic derived from historical traffic patterns. Although effective under moderate demand, these approaches struggle in highly dynamic environments characterized by stochastic flows and nonlinear interactions. Consequently, recent research has increasingly explored Deep Reinforcement Learning (DRL) as a data-driven alternative for adaptive traffic signal control \cite{2,10}.
    In DRL-based traffic control, an intersection is modeled as an agent that learns signal policies through interaction with its environment, as illustrated in Fig.~\ref{fig:component}. Methods based on Deep Q-Networks, policy-gradient algorithms, and multi-agent learning have demonstrated the ability to reduce delay and queue length compared to conventional controllers \cite{3, 7, 14}. Despite these advances, one critical design element remains insufficiently understood: the reward function. Because the reward defines the optimization objective, it directly shapes learned policies. Most existing formulations rely on delay or queue-based penalties, which can induce undesirable behaviors such as excessive phase switching \cite{12, 16}. Such flickering policies may minimize instantaneous congestion but disrupt vehicle platoons, producing stop-and-go oscillations that increase fuel consumption, emissions, and collision risk \cite{8, 9, 15}.
    Current mitigation strategies typically impose fixed constraints, such as minimum green times or switching penalties. While these heuristics improve stability, they reduce responsiveness because they do not adapt to real-time traffic conditions. This reveals a fundamental limitation of conventional reward design: the objective often penalizes congestion after it forms rather than encouraging stable flow dynamics. In this context, PressLight \cite{press} integrates Max Pressure \cite{maxp} as a reward function. However, this approach exhibits a single-objective bias, optimizing throughput without considering other performance metrics.
    To address this limitation, we propose a Momentum-Based Reward Function (MBRF) that incorporates traffic flow continuity directly into the learning objective. Instead of penalizing delay, the reward promotes sustained vehicle motion by incentivizing phase persistence proportional to observed discharge efficiency. This formulation aligns the learning objective with physical traffic behavior, encouraging smoother control policies without rigid constraints.
    
    This work introduces a momentum-based reward for traffic signal control that promotes continuous vehicle movement. Results show improved traffic efficiency and lower emissions without explicitly optimizing environmental metrics.


    \section{Related Work}
    \label{rel:related_work}
    Traffic signal control has evolved from analytical timing methods to adaptive and learning-based strategies. Early approaches derived optimal signal timings using traffic flow theory, while later adaptive systems introduced feedback mechanisms to adjust parameters in real time. Network-level controllers, such as max-pressure methods, extended these ideas by using queue dynamics to achieve stability guarantees. Although theoretically grounded, these methods depend on calibrated models and may degrade under highly variable traffic conditions.
    Reinforcement learning offers a flexible alternative that learns control policies directly from interaction data. Early studies showed that RL controllers can outperform fixed-time signals, and subsequent work incorporated deep neural networks to handle high-dimensional traffic states \cite{ 7, 14, 4}. Recent research has focused on scalability through decentralized and multi-agent frameworks capable of coordinating large networks \cite{10}. These advances demonstrate the potential of DRL but also reveal sensitivity to system design choices, particularly state representation and reward formulation \cite{2}.
    Reward design is especially influential because it determines the objective optimized during training. Most existing work uses penalty-based rewards derived from delay, queue length, travel time, or weighted combinations of performance metrics \cite{ 3, 12, 13, 5, 20}. Although effective in some scenarios, such rewards may provide sparse feedback or bias optimization toward a single metric. Several studies aim to stabilize learning by using switching penalties, action masking, or explicit platoon detection, but these approaches either constrain policy flexibility or introduce additional modeling complexity \cite{16, 20}.
    Recent efforts have incorporated environmental objectives by embedding emission estimation models into the reward signal \cite{9}. While this can improve sustainability metrics, emission estimates are often noisy, and using them can be computationally expensive. Another work \cite{agand} proposed weighted reward-shaping schemes to reduce CO2 emissions. However, it is important to note that their approach depends on the types of vehicles, where they assume the aid of a computer vision system to classify vehicle types and then optimize traffic to reduce CO2 emissions.
    An alternative is to promote stable traffic flow directly, since instability is a primary cause of excess emissions \cite{8, 15}.
    Motivated by these observations, this work investigates a motion-centric reward that encourages continuous vehicle movement rather than penalizing congestion. The proposed formulation is compatible with different traffic scenarios and can be extended to heterogeneous traffic settings. By aligning the learning objective with traffic flow dynamics, the proposed approach promotes stable policies that improve efficiency, robustness, and environmental performance.

    \section{Problem Formulation}
    
    Adaptive traffic signal control is formulated as a Markov Decision Process (MDP), a standard framework used in reinforcement learning-based traffic control. The problem is defined by the tuple $(\mathcal{S}, \mathcal{A}, \mathcal{P}, R, \gamma)$, where $\mathcal{S}$ denotes the state space, $\mathcal{A}$ the action space, $P$ the transition dynamics, $R$ the reward function, and $\gamma \in [0,1]$ the discount factor. At each decision step, the agent observes the traffic state, selects a signal phase, and receives a reward determined by the resulting traffic conditions.
    
    \subsection{State Representation}
    

    \begin{figure}[t]
    \centering
    \includegraphics[width=0.6\linewidth]{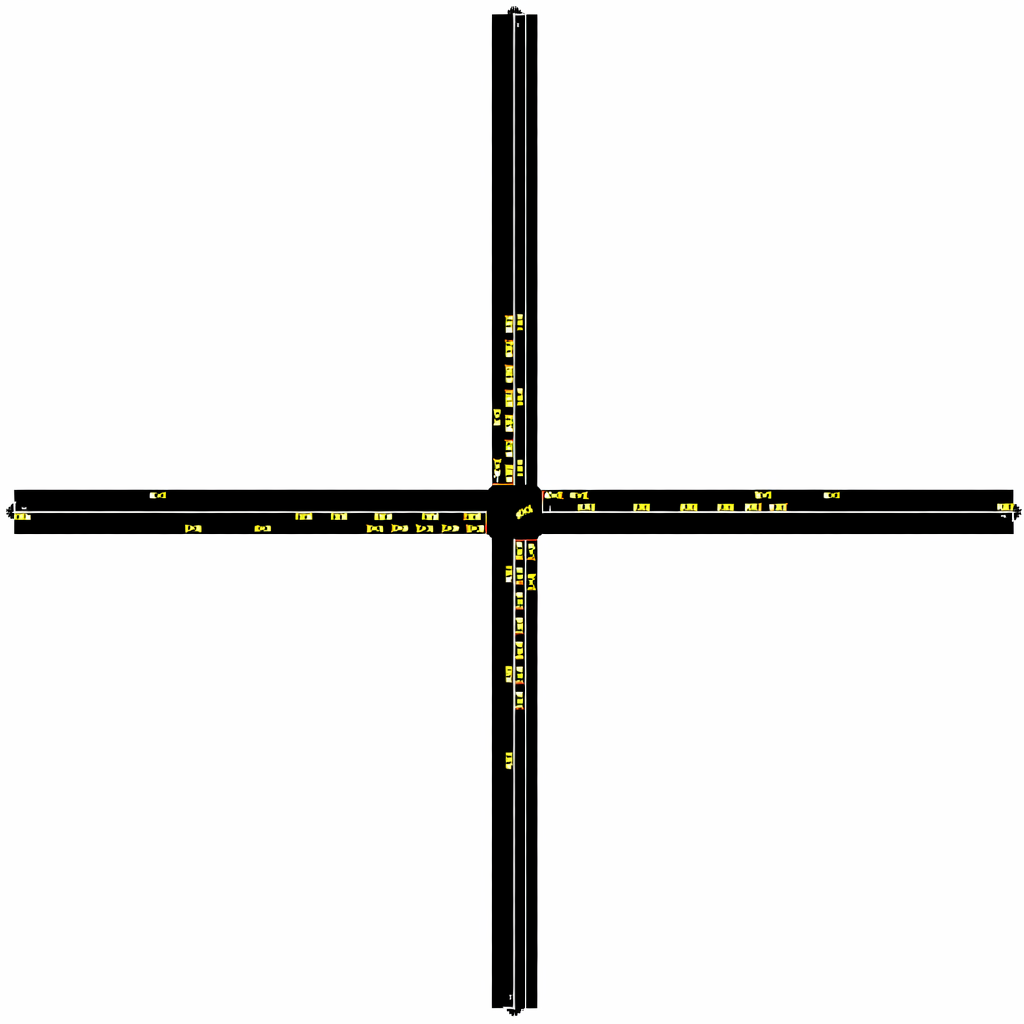}
    \caption{Two-way single intersection with through, left, and right options in each lane with different road users}
    \label{fig:int}
    \vspace{-0.2cm}
    \end{figure}
        
    The state vector captures both traffic conditions and signal status, following commonly used representations in DRL-based traffic signal control \cite{agand, sumorl}:
    \begin{equation}
    \mathbf{s} = [p,\; g,\; d_1,\dots,d_n,\; q_1,\dots,q_n],
    \end{equation}
    where $p$ denotes the current signal phase, $g$ indicates whether the minimum green time constraint has been satisfied, $d_i$ represents lane density, and $q_i$ denotes queue length for lane $i$. Such hybrid state representations combining signal information and traffic measurements have been shown to provide sufficient information for learning effective signal policies \cite{sumorl}. Fig.~\ref{fig:int} shows the intersection environment.
    
    

    \subsection{Action Space}
    
    The action space is discrete and determines which green phase is activated next. For the two-phase intersection used in this study, the agent selects between North–South green or East–West green, giving $\mathcal{A}=\{0,1\}$. Each action applies the chosen phase for a fixed control interval $\Delta t$. If the selected phase differs from the current one, the simulator automatically inserts a yellow transition phase before switching.

    \subsection{Reward Design}
    
    Reward formulation is critical because it defines the learning objective and strongly influences policy behavior. Conventional approaches typically use delay or queue-based penalties to minimize congestion. While effective in some cases, such rewards may encourage short-term optimization and unstable switching behavior.
    
    To address this limitation, we propose a \textbf{Momentum-Based Reward Function (MBRF)} that encourages sustained traffic flow. The reward at time $t$ is defined as
    \begin{equation}
    R_t = \frac{1}{M} \sum_{i=1}^{M} m_i v_i,
    \end{equation}
    where $m_i$ is the mass of vehicle, $v_i$ is the speed of vehicle $i$, and $M$ is the number of vehicles in the intersection region. In the homogeneous setting ($m_i=1$), this formulation effectively incentivizes sustained vehicle motion through a speed-based objective, promoting smoother traffic flow and reducing stop-and-go behavior, which are known contributors to increased emissions and instability \cite{10}. In the heterogeneous setting with mixed traffic, we assign different $m_i$ values to vehicles such as buses, trucks, and bikes, thereby creating a true momentum-based metric. This penalizes the deceleration of high-mass vehicles more strongly and helps reduce $CO_2$ emissions.


    We compare the proposed reward with commonly used formulations: waiting-time reward \cite{wait_rew}, \( r_t = -w_t \); queue-length reward \cite{sumorl}, \( r_t = -q_t \); and differential waiting reward \cite{3}, \( r_t = w_{t-1} - w_t \).
    
    \subsection{Learning Algorithm}
    
    We employ a Deep Q-Network (DQN) to learn optimal traffic signal control policies. In reinforcement learning, an agent interacts with the environment by observing a state $\textbf{s}_t$, selecting an action $a_t$, receiving a reward $R_{t+1}$, and transitioning to a new state $\textbf{s}_{t+1}$. The objective is to maximize the expected discounted return~\cite{Sutton1998}
    \begin{equation}
    G_t = \sum_{k=0}^{\infty} \gamma^k R_{t+k+1},
    \end{equation}
    where $\gamma \in (0,1]$ is the discount factor.
    
    To balance exploration and exploitation, actions are selected using an $\epsilon$-greedy policy:
    \begin{equation}
    a_t =
    \begin{cases}
    \text{random action}, & \text{with probability } \epsilon,\\
    \argmax\limits_{a} Q(\textbf{s}_t,a), & \text{with probability } 1-\epsilon.
    \end{cases}
    \end{equation}
    The Q-values are updated using
    \begin{equation}
    Q(\textbf{s}_t,a_t) \leftarrow Q(\textbf{s}_t,a_t) +
    \alpha \Big[ R_{t+1} + \gamma \max_a Q(\textbf{s}_{t+1},a) - Q(\textbf{s}_t,a_t) \Big],
    \end{equation}
    where $\alpha$ is the learning rate. In DQN, a neural network approximates $Q(\textbf{s},a)$ for all actions, enabling scalable learning in high-dimensional state spaces.
    
    To stabilize training, we use experience replay and a separate target network. The target network is a copy of the main Q-network and is used to compute the target value in the update step, while the main network is trained. The target network is periodically updated with the weights of the main network, which helps reduce instability and prevents the learning process from diverging. The Q-network is implemented as a multilayer perceptron with two hidden layers of 64 neurons each and ReLU activations. The agent learns a policy $\pi(a|\textbf{s})$ that maximizes the expected cumulative reward
    \begin{equation}
    J(\pi) = \mathbb{E}\Big[\sum_{t=0}^{\infty} \gamma^t R_t \Big],
    \end{equation}
    which encourages sustained vehicle motion and improved traffic flow efficiency. 
    
    \begin{table}[t]
    \centering
    \caption{DQN Hyperparameters for Traffic Signal Control}
    \label{tab:dqn_params}
    \begin{tabular}{lc}
    \hline
    \textbf{Parameter} & \textbf{Description / Value} \\
    \hline
    m & 1 \\
    Learning Rate ($\alpha$) & 0.001 \\
    Discount Factor (\(\gamma\)) & 0.99 \\
    Target Network Update Interval & 500 steps \\
    Exploration (\(\epsilon\)-greedy) & 0.05 → 0.01 \\
    Replay Buffer & Enabled \\
    Episode Duration & 1000 sec per episode \\
    Evaluation Frequency & Every 10,000 steps \\
    \hline
    \end{tabular}
    \vspace{-0.2cm}
    \end{table}
    
    \section{Experimental Setup and Evaluation}
    \label{sec:experiments}
    
    \subsection{Simulation Environment}
    
    Experiments are conducted using the SUMO (Simulation of Urban MObility) traffic simulator. The environment consists of a four-arm single intersection with two-phase control (North--South and East--West). Traffic demand is generated using randomized route files to create variable density conditions and avoid overfitting to a single traffic pattern.
    
    Each episode corresponds to a complete simulation rollout. To ensure statistical reliability, results are averaged over multiple independent evaluation episodes. Signal timing parameters follow fixed operational limits: minimum green time $g_{\min}=5$\,s, maximum green time $g_{\max}=50$\,s, yellow phase duration of 2\,s, and control interval $\Delta t=5$\,s. These values match the simulator configuration used during training and evaluation.
    
    \subsection{Training Configuration}
    
    The DQN agent is trained for a fixed number of interaction steps using experience replay and a target network for stabilization. Traffic demand patterns are randomized across episodes to improve generalization. The network hyperparameters used in training are summarized in Table~\ref{tab:dqn_params}, and all reward formulations are trained under identical hyperparameters to ensure fair comparison. 
    
    \subsection{Baseline Controllers}
    
    We compare the proposed Momentum-Based Reward Function (MBRF) against learning-based baselines using waiting-time, queue-length, and differential waiting rewards, as well as classical controllers including Max Pressure and Longest Queue First (LQF). All reinforcement learning agents use the same network architecture and training configuration to ensure fair comparison.

    \subsection{Evaluation Metrics}
    
    Performance is evaluated using standard traffic engineering metrics:
    
    \begin{itemize}
        \item \textbf{Average Waiting Time} (s)
        \item \textbf{Average Queue Length}
        \item \textbf{Throughput} (vehicles completed)
        \item \textbf{Average Travel Time} (s)
        \item \textbf{Total CO$_2$ Emissions}
    \end{itemize}
    
    CO$_2$ emissions are computed directly from SUMO’s emission model. These metrics are widely used in traffic signal control studies \cite{20}.
    
    \subsection{Statistical Evaluation}
    
    To account for stochasticity in traffic generation and learning, each experiment is repeated across 3 random seeds. Reported results correspond to mean performance, and standard deviation is included where appropriate.

    \section{Results and Discussion}
    \label{sec:results}

    \begin{table*}[t]
    \centering
    \caption{Performance averaged across three random seeds. Best values per metric are shown in bold.}
    \label{tab:results}
    \begin{tabular}{lccccc}
    \hline
    Method & Waiting (s) & Queue & Throughput & Travel (s) & CO$_2$ (g) \\
    \hline
    DQN-Wait & 320.1 $\pm$ 0.2 & 79.8 $\pm$ 0.0 & 381 $\pm$ 1 & \textbf{22.6 $\pm$ 0.2} & 28362.1 $\pm$ 78.7 \\
    DQN-Queue & 9.3 $\pm$ 3.6 & 13.6 $\pm$ 2.1 & 654 $\pm$ 5 & 49.5 $\pm$ 3.7 & 16818.4 $\pm$ 1040.4 \\
    DQN-Diff & \textbf{4.4 $\pm$ 0.5} & 13.6 $\pm$ 2.5 & 656 $\pm$ 9 & 47.8 $\pm$ 5.1 & 16219.6 $\pm$ 1776.7 \\
    \textbf{Proposed (MBRF)} & 16.7 $\pm$ 11.0 & \textbf{13.0 $\pm$ 2.6} & \textbf{664 $\pm$ 4} & 41.9 $\pm$ 1.2 & \textbf{14337.1 $\pm$ 673.5} \\
    Max Pressure & 323.6 $\pm$ 77.5 & 58.1 $\pm$ 10.9 & 348 $\pm$ 201 & 26.5 $\pm$ 2.0 & 22437.0 $\pm$ 406.5 \\
    Longest Queue & 314.1 $\pm$ 95.5 & 64.4 $\pm$ 5.7 & 323 $\pm$ 196 & 35.7 $\pm$ 7.8 & 24486.5 $\pm$ 1611.7 \\
    \hline
    \end{tabular}
    \end{table*}
    \begin{figure}[t]
    \centering
    \includegraphics[width=\linewidth]{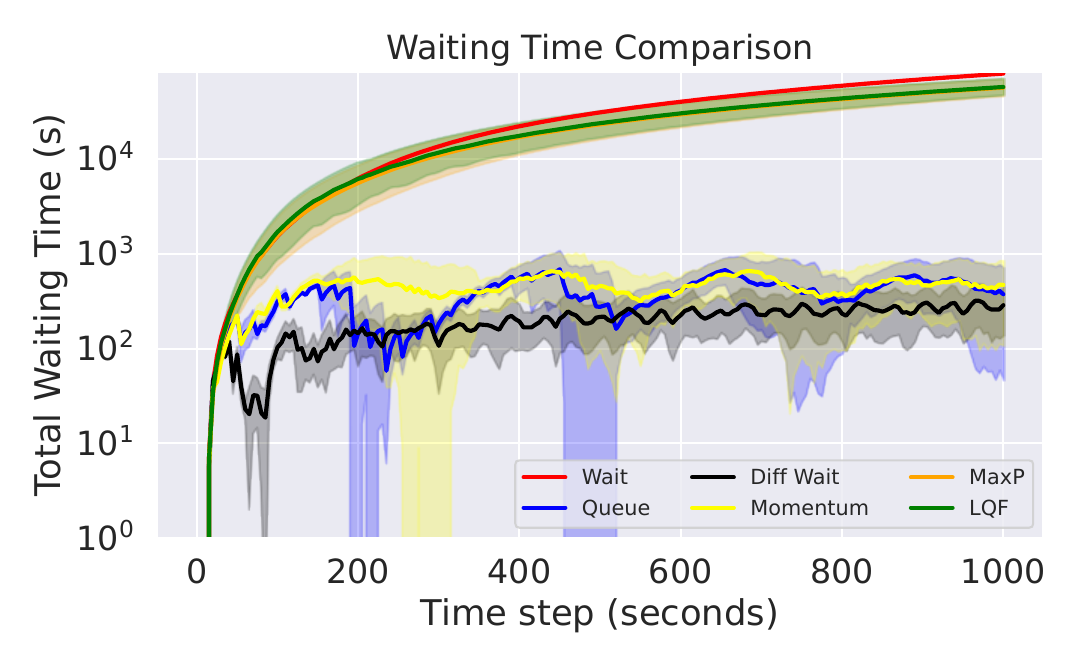}
    \caption{Average waiting time during evaluation episodes.}
    \label{fig:waiting}
    \end{figure}
    
    \begin{figure}[t]
    \centering
    \includegraphics[width=\linewidth]{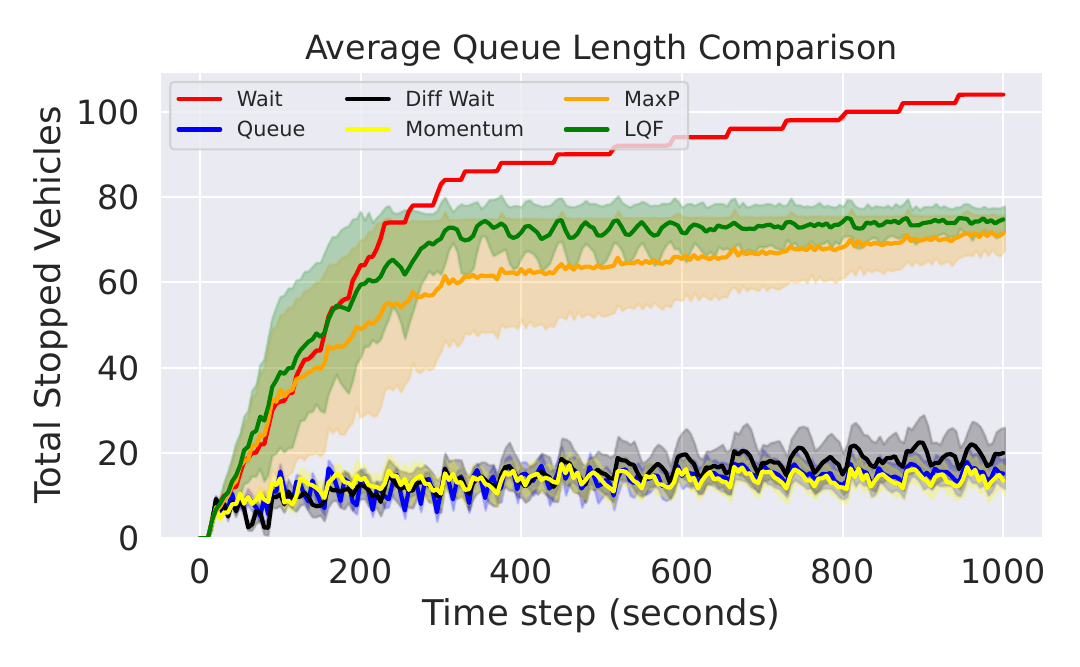}
    \caption{Average queue length during evaluation episodes.}
    \label{fig:queue}
    \end{figure}
    
    \begin{figure}[t]
    \centering
    \includegraphics[width=\linewidth]{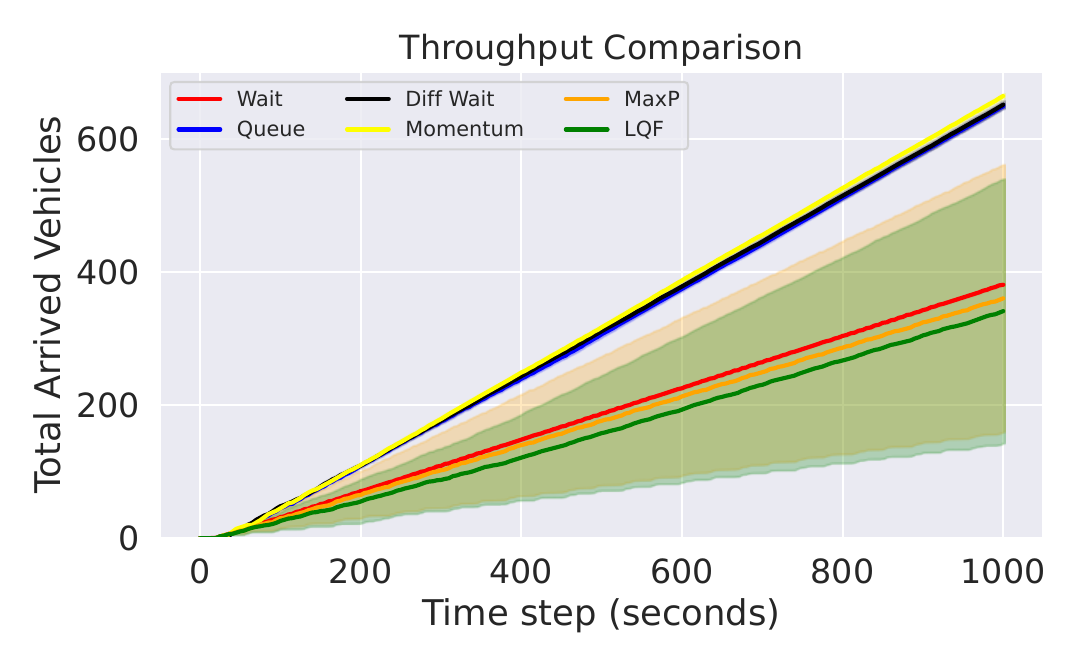}
    \caption{Throughput comparison across methods.}
    \label{fig:throughput}
    \end{figure}
    
    \begin{figure}[t]
    \centering
    \includegraphics[width=\linewidth]{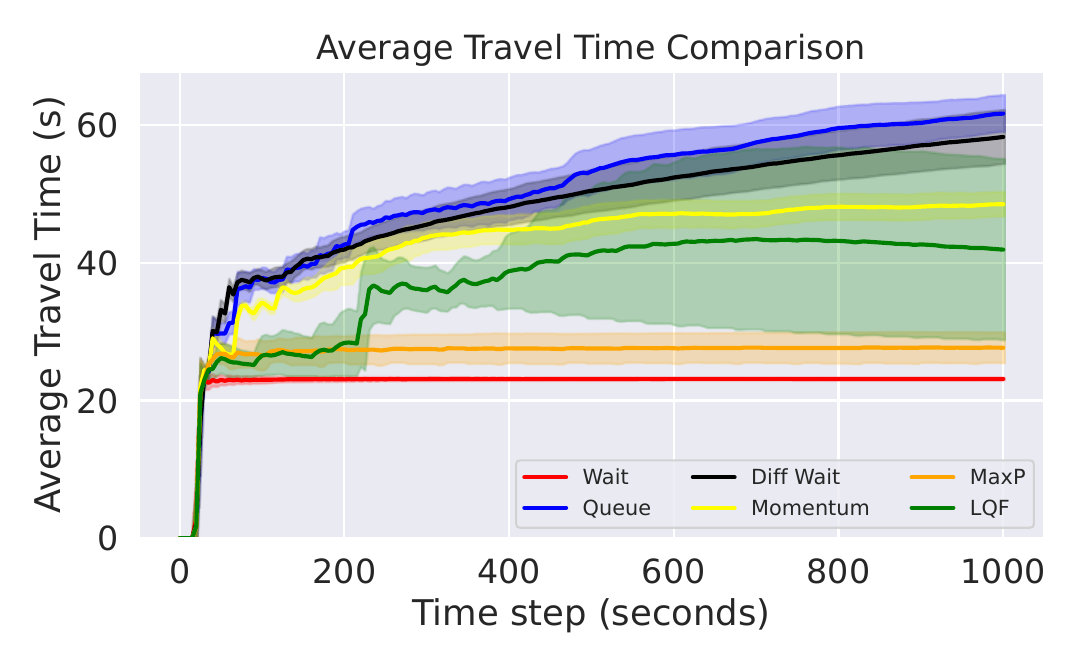}
    \caption{Average travel time comparison.}
    \label{fig:travel}
    \end{figure}
    
    \begin{figure}[t]
    \centering
    \includegraphics[width=\linewidth]{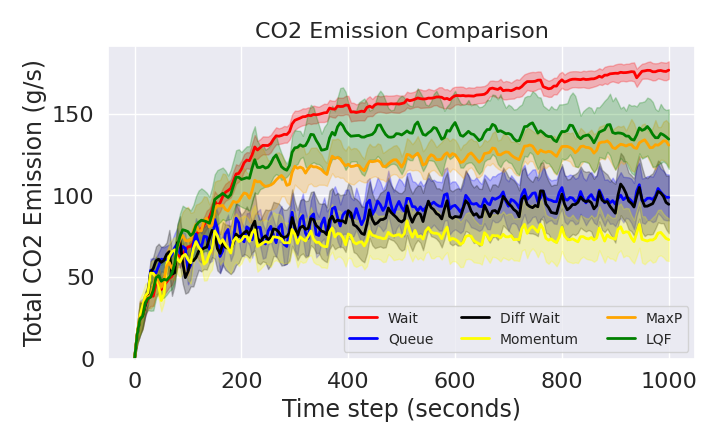}
    \caption{Total CO$_2$ emissions comparison.}
    \label{fig:co2}
    \end{figure}

    Initially, we show the results for the homogeneous setting in sections \ref{sec:Traffic} to \ref{sec:Training}. Table~\ref{tab:results} summarizes the quantitative performance comparison across all evaluated controllers, and Figs.~\ref{fig:waiting}--\ref{fig:co2} illustrate the temporal evolution of each metric. Reported values correspond to the mean $\pm$ standard deviation across three random seeds. Then we discuss the heterogeneous setting in Sec.~\ref {sec:hetero} and the results are summarized in Table \ref{tab:hetero_results}.
    
    \subsection{Traffic Efficiency and Congestion Dynamics}\label{sec:Traffic}
    
    Fig.~\ref{fig:waiting} shows that the waiting-time reward baseline produces rapidly increasing cumulative waiting, consistent with its high final value of 320.1\,s in Table~\ref{tab:results}. Classical controllers exhibit similarly large totals (Max Pressure: 323.6\,s, Longest Queue First: 314.1\,s), indicating limited congestion mitigation capability under the tested demand conditions. In contrast, the differential waiting reward achieves the lowest mean waiting time (4.4\,s), followed by the queue-based reward (9.3\,s). The proposed momentum-based reward maintains a moderate waiting time of 16.7\,s, remaining significantly lower than classical baselines while avoiding the extreme growth observed in delay-minimization approaches.
    
    Queue dynamics in Fig.~\ref{fig:queue} further highlight these differences. The waiting-time reward leads to severe queue accumulation (79.8 vehicles on average), and classical controllers also sustain large queues (58.1 and 64.4 vehicles for Max Pressure and LQF, respectively). The proposed method achieves the lowest mean queue length (13.0 vehicles), slightly outperforming both the queue-based and differential-reward methods (13.6 vehicles each). Additionally, the queue trajectory under the momentum reward stabilizes earlier and exhibits smaller variance, suggesting faster convergence to steady operating conditions.
    
    Throughput trends shown in Fig.~\ref{fig:throughput} demonstrate clear separation between classical and learning-based strategies. Classical controllers terminate at substantially lower totals (348 and 323 vehicles), whereas learning-based approaches maintain near-linear growth throughout the episode. The proposed method achieves the highest throughput overall (664 vehicles), exceeding differential waiting (656 vehicles) and queue-based reward (654 vehicles). This indicates that encouraging sustained vehicle motion improves effective intersection utilization beyond what is achieved by directly penalizing congestion measures.

    \begin{table*}[t]
    \centering
    \caption{Heterogeneous traffic control. Best values per metric are shown in bold.}
    \label{tab:hetero_results}
    \begin{tabular}{lccccc}
    \hline
    Method & Waiting (s) & Queue & Throughput & Travel (s) & CO$_2$ (g) \\
    \hline
    DQN-Wait & 296.2 $\pm$ 4.4 & 57.8 $\pm$ 2.3 & 231 $\pm$ 3 & 62.3 $\pm$ 3.3 & 20571.3 $\pm$ 722.9 \\
    DQN-Queue & 16.1 $\pm$ 11.7 & 22.1 $\pm$ 13.8 & 429 $\pm$ 60 & 65.0 $\pm$ 15.4 & 15257.9 $\pm$ 6722.6 \\
    DQN-Diff & \textbf{10.5 $\pm$ 13.1} & \textbf{15.1 $\pm$ 12.7} & 455 $\pm$ 61 & 53.9 $\pm$ 13.8 & 12808.6 $\pm$ 6742.3 \\
    \textbf{Proposed (MBRF)} & 11.8 $\pm$ 7.0 & 15.2 $\pm$ 10.1 & \textbf{519 $\pm$ 108} & 49.6 $\pm$ 8.6 & \textbf{12545.0 $\pm$ 5310.9} \\
    Max Pressure & 298.4 $\pm$ 100.1 & 52.3 $\pm$ 6.7 & 348 $\pm$ 193 & \textbf{36.6 $\pm$ 9.0} & 22052.7 $\pm$ 2345.2 \\
    Longest Queue & 275.7 $\pm$ 102.7 & 57.2 $\pm$ 5.1 & 356 $\pm$ 188 & 44.3 $\pm$ 6.9 & 23632.1 $\pm$ 2767.4 \\
    \hline
    \end{tabular}
    \vspace{-0.25cm}
    \end{table*}

    \subsection{Travel Time Performance}
    
    Average travel time results further illustrate trade-offs between objectives. Although the waiting-time reward achieves the lowest travel time (22.6\,s), it is accompanied by extremely poor queue and emission performance. Among the more balanced approaches, the proposed momentum-based reward achieves an average travel time of 41.9\,s, improving upon queue-based (49.5\,s) and differential waiting (47.8\,s) rewards. This suggests that promoting sustained motion enhances end-to-end traffic efficiency rather than merely minimizing instantaneous delay.
    
    \subsection{Emission Performance}
    
    CO$_2$ emission trends in Fig.~\ref{fig:co2} closely follow stability and throughput patterns. The proposed method achieves the lowest total emissions (14337\,g), outperforming queue-based (16818\,g) and differential waiting (16219\,g) rewards. Compared to the waiting-time reward (28362\,g), emissions are reduced by approximately 49\%. Classical baselines also produce substantially higher emissions (22437\,g for Max Pressure and 24486\,g for LQF). These results support the hypothesis that smoother traffic flow reduces stop-and-go oscillations, thereby lowering fuel consumption and emissions without explicitly optimizing for environmental metrics.
    
    \subsection{Training Consistency and Performance}\label{sec:Training}
    
    Across all figures, delay-oriented rewards exhibit wider confidence intervals, indicating higher variability under stochastic traffic demand. In contrast, the proposed reward demonstrates lower variance across random seeds, particularly in throughput ($\pm$4 vehicles) and emissions ($\pm$673\,g). The relatively small standard deviations suggest consistent policy performance and reduced sensitivity to initialization.
    
    
    
    Overall, these results suggest that the motion-centric reward provides a well-conditioned learning signal, leading to consistent training outcomes while maintaining responsiveness to traffic dynamics.

    \subsection{Heterogeneous Traffic}\label{sec:hetero}
    
    While the primary evaluation utilizes a uniform mass ($m=1$) to isolate general motion dynamics, real-world traffic is highly heterogeneous. Hence, we conducted an evaluation incorporating actual vehicle masses queried directly from the simulator with a distribution of 50\% passenger cars, 20\% trucks, 15\% buses, and 15\% motorcycles. These classes are assigned masses of 1,500 kg, 8,000 kg, 12,000 kg, and 200 kg, respectively. The MBRF calculates rewards by scaling each vehicle's instantaneous momentum relative to a standard car mass (1,500 kg). Under this formulation, reductions in the speed of high-mass vehicles contribute more strongly to the reward signal than equivalent reductions for passenger cars.

    As summarized in Table~\ref{tab:hetero_results}, this heterogeneous momentum reward achieved a throughput of 519 vehicles and reduced total CO$_2$ emissions to 12545\,g during a single-seed run. The differential waiting reward yielded a slightly lower average queue length (15.1 $\pm$ 12.7) than the proposed MBRF (15.2 $\pm$ 10.1). However, the difference is very small relative to the variance. In contrast, the proposed method showed more consistent performance across evaluation metrics while maintaining the highest throughput and lowest emissions. Max Pressure achieves a lower average travel time in this setting, but it performs worse in the remaining metrics, including waiting time, queue length, throughput, and CO$_2$ emissions.




    \section{Conclusion and Future Work}
    \label{sec:conclusion_future}
    
    In conclusion, the results demonstrate that optimizing a single congestion metric leads to clear performance trade-offs. While other rewards reduce nominal travel time, these approaches either degrade throughput, increase emissions, or exhibit instability. Classical pressure-based heuristics remain stable but underutilize intersection capacity. Among the evaluated methods, the proposed reward achieves the highest throughput, the lowest queue length, the lowest emissions, and competitive travel time. These findings indicate that directly incentivizing vehicle motion enables implicit multi-objective optimization, yielding balanced improvements in efficiency, stability, and environmental performance without requiring manually weighted reward terms in the evaluated intersection setting.
    
A notable limitation of this work is that experiments were performed solely on a fully observable, isolated intersection with basic two-phase control. Deployed traffic systems are inherently more complex, frequently contending with limited sensor visibility, communication latency, and broader multi-intersection networks. Currently, experiments have been conducted only with the DQN algorithm. Future work will involve evaluating the method with other RL approaches, such as PPO and actor–critic methods, and implementing larger multi-intersection traffic networks.
	
	\bibliographystyle{IEEEtran}
	\bibliography{root} 
	
\end{document}